\pdfoutput=1
\documentclass[a4paper, 10pt, conference]{ieeeconf}      % Use this line for a4 paper

\IEEEoverridecommandlockouts                              % This command is only needed if 
\usepackage{graphicx}
\usepackage{hhline}
\usepackage{amsmath}
\usepackage{booktabs, makecell, multirow, rotating}
\usepackage{float}
\usepackage{caption}
\usepackage{cite}
\usepackage{textcomp}
\usepackage{subfigure}
\usepackage{multicol, blindtext}
\usepackage{tabularx}
\usepackage{hyperref}
\usepackage[dvipsnames]{xcolor}

\let\labelindent\relax
\usepackage{enumitem}
\usepackage{gensymb}
\usepackage{algorithm}
\usepackage{algorithmic}
\usepackage{microtype} % recommended
\usepackage{colortbl}
\usepackage{amssymb}

\title{\LARGE \bf

DexTwist: Dexterous Hand Retargeting for Twist Motion via Mixed Reality-based Teleoperation
}

\author{Dongmyoung Lee$^{*1}$, Chengxi Li$^{*1}$, and Dongheui Lee$^{1, 2}$% <-this % stops a space
\thanks{$^{1}$Autonomous Systems Lab, Institute of Computer Technology, TU Wien, Vienna, Austria.}
\thanks{$^{2}$Institute of Robotics and Mechatronics (DLR), German Aerospace Center, Wessling, Germany.}
\thanks{* Authors contributed equally}
\thanks{This work has been partially supported by the European Union project INVERSE under grant agreement No. 101136067.}
}

\begin{document}
\maketitle
\begin{abstract}
Dexterous teleoperation via Mixed Reality (MR)-based interfaces offers a scalable paradigm for transferring human manipulation skills to dexterous robot hands. However, conventional retargeting approaches that minimize kinematic dissimilarity (e.g., joint angle or fingertip position error) often fail in contact-rich rotational manipulation, such as cap opening, key turning, and bolt screwing. This failure stems from the embodiment gap: mismatched link lengths, joint axes/limits, and fingertip geometry can cause direct pose imitation to induce tangential fingertip sliding rather than stable object rotation, resulting in screw axis drift, contact slip, and grasp instability. To address this, we propose DexTwist, a functional twist-retargeting framework for MR-based dexterous teleoperation. DexTwist detects a tripod pinch, estimates the operator's intended screw axis and twist magnitude, and applies a real-time residual joint-space refinement that tracks turning progress while regularizing the robot tripod geometry. The refinement minimizes a virtual-object objective defined by turning angle, screw axis consistency, fingertip closure, and tripod stability. Simulation and real-world experiments show that DexTwist improves turning angle tracking and screw axis stability compared with a vector-based retargeting baseline.
\end{abstract}

\section{Introduction}

Dexterous robotic hands offer the promise of human-level manipulation capability, but reliable dexterous control remains challenging because contact-rich tasks require precise coordination of high-dimensional end-effectors to interact with diverse objects. Recent advances in vision-based hand tracking allow operators to transfer manipulation skills to robotic hands without specialized master devices or motion-capture systems, lowering deployment barriers in real-world environments~\cite{handa2020dexpilot,qin2023anyteleop,fang2017novel}. Meanwhile, building on these advances, Mixed-Reality Head-Mounted Display (MR-HMD) provides intuitive and mobile teleoperation interfaces for robot learning demonstrations and human fallback when autonomous execution fails~\cite{cheng2024open,ding2025bunny}. Throughout this paper, MR-based teleoperation leverages a headset interface that combines egocentric hand tracking and streamed visual feedback from the robot workspace. This interface allows the operator to provide hand/wrist motion commands while monitoring the real-world manipulation scene through the headset, forming an integrated sensing and feedback platform for dexterous teleoperation.

In terms of retargeting, despite high fidelity in free-space imitation, teleoperation often struggles in contact-rich interactions such as opening a cap, rotating a knob, or twisting a tool. A primary bottleneck is the embodiment gap: differences in kinematics and geometry between human and robotic hands mean that joint- or pose-level similarity does not necessarily produce correct contact behavior. Conventional retargeting objectives therefore work well for gesture imitation but provide weak guarantees for contact-rich manipulation~\cite{Antotsiou_2018_ECCV_Workshops, li2019vision}.

For rotational tasks such as screwing, pose similarity is an insufficient proxy for task success. The robot should generate rotation about a physical axis while maintaining stable fingertip contacts and avoiding slip. Under embodiment mismatch, direct kinematic mapping can produce tangential fingertip motion along the object surface rather than rotation about the intended axis, leading to screw axis drift, contact slip, and loss of grasp stability. These failure modes motivate a shift from gesture imitation to function preservation.

\begin{figure}[t!]
    \centering
    \includegraphics[width=0.98\columnwidth]{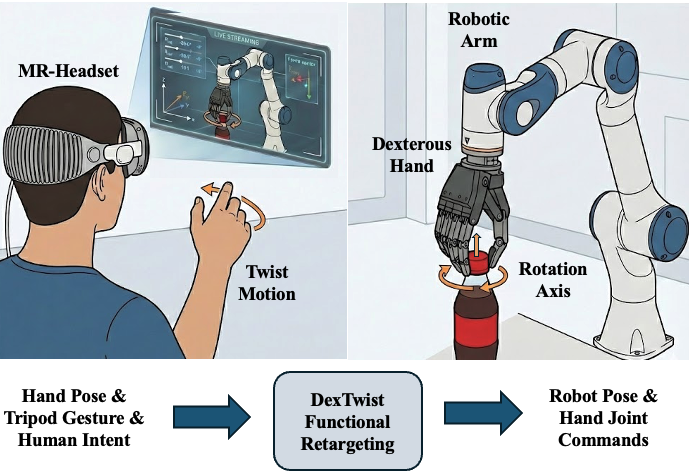}
    \caption{An overview of the proposed system DexTwist, a robot teleoperation and hand retargeting framework that supports achieving dexterous screw manipulation}
    \label{fig:concept_overview}
    \vspace{-10pt}
\end{figure}

In this work, we propose DexTwist, an MR-based teleoperation system with functional hand retargeting for in-hand twist manipulation, as shown in Fig.~\ref{fig:concept_overview}. DexTwist commands the robot arm, detects a human tripod pinch, estimates the screw axis and accumulated twist intent in the operator's palm frame, and generates robot hand commands through real-time residual refinement. The objective tracks the intended turning angle while penalizing screw axis drift, loss of fingertip closure, and tripod centroid translation, preserving the turning function under human-robot embodiment mismatch.

Reliable twist teleoperation is relevant to remote maintenance, inspection, disaster response, and industrial servicing, where operators may need to manipulate caps, keys, valves, or fasteners without direct physical presence. The contributions of this work are summarized as follows:

\begin{enumerate}[leftmargin=*]
\item An MR-based dexterous teleoperation pipeline with robust tripod activation and episode-level turning state that supports ratcheting.
\item A functional twist retargeting module implemented as a residual joint-space refinement, minimizing a virtual-object objective defined in the robot palm frame to track turning progress while regularizing tripod stability.
\item An evaluation protocol comparing DexTwist with vector-based retargeting in simulation and real-world hardware.
\end{enumerate}

\section{Related Works} \label{sec:relatedworks}
\subsection{Robot Teleoperation}

Teleoperation has become a key paradigm for large-scale robotic manipulation data collection, evolving from parallel-jaw grippers \cite{seo2023deep,fu2024mobile} to anthropomorphic dexterous hands. While multi-fingered platforms enable richer manipulation capabilities, they substantially increase teleoperation complexity, leading prior work to explore exoskeletons \cite{fang2024airexo}, motion capture systems \cite{liu2019high}, MR devices \cite{arunachalam2022holo}, and vision-based interfaces. 

Among these, MR-based approaches offer improved accessibility and immersion, yet existing systems largely focus on pose-level control and data acquisition, leaving functional manipulation intent and contact-aware execution underexplored. Lipton et al. \cite{lipton2017baxter} and Wan et al. \cite{wan2024virtual} leverage off-the-shelf Virtual-Reality(VR) hardware to enable immersive human control, they achieve this by embedding the operator in a virtual control room mapped to the robot’s workspace while providing rich sensor feedback and dynamic human–robot mapping. Ponomareva et al.\cite{ponomareva2021grasplook} proposed an augmented virtual environment for medical teleoperation that alleviates the limitations of camera-based visual feedback, such as restricted viewpoints and high cognitive load. Regarding the MR devices, Ding et al. introduce Bunny-VisionPro \cite{ding2025bunny}, a real-time bimanual dexterous teleoperation system that leverages an MR headset, and further design novel low-cost devices to provide haptic feedback to the operator, enhancing immersion. Similarly, Cheng et al. \cite{cheng2024open} presented an immersive teleoperation system designed to provide stereoscopic perception and mirroring the operator’s arm and hand motions onto humanoid robots. Moreover, some recent works also focus on scalable, user-friendly teleoperation platforms to address the scarcity of high-quality robotic manipulation data. OpenTeach\cite{iyer2024open} enables low-cost, immersive MR-based teleoperation across diverse robot morphologies for collecting dexterous, contact-rich demonstrations, while DART \cite{park2024dexhub} reimagines data collection through cloud-based simulation and MR devices, enabling crowdsourced, high-throughput data generation with reduced operator fatigue. Both systems demonstrate that data collected through intuitive teleoperation—either in real-world MR or cloud simulation—can effectively support policy learning and transfer to real robots.

Previous research has highlighted the effectiveness of MR applications in the realm of teleoperation with ease. Currently, the predominant focus of utilizing MR-headset revolves around extracting human information for robot motion control, it still does not leverage the rich visualization feature well for providing information for human operator to visualization and monitoring.

\subsection{Hand Motion Retargeting}

In dexterous hand teleoperation, accurately retargeting human hand motion to robotic hands remains challenging because human and robotic hands differ in morphology, kinematic structure, joint limits, fingertip geometry, and degrees of freedom. Prior work has addressed this problem using hand motion captured from motion-capture systems, VR headsets, data gloves, or vision-based tracking. Learning-based methods align human and robot hand motions through shared latent spaces or directly map visual hand observations to robot joint commands~\cite{yan2025learning,li2019vision}. These approaches are attractive for scalable retargeting, but often require representative training data and tend to reproduce hand pose or joint configurations. Optimization-based methods instead formulate retargeting through fingertip positions, task-space vectors, contact constraints, or inverse kinematics~\cite{lakshmipathy2025kinematic,orbik2021human,handa2020dexpilot}. Such methods improve transfer across different embodiments by explicitly encoding geometric correspondences, but their objectives are still largely based on pose or contact geometry. As a result, they may not preserve the intended manipulation function during contact-rich twisting, where task success depends on stable rotation and torque transmission rather than hand-shape similarity alone.

Recent work has explored functional retargeting by shifting from hand-pose imitation to object-centric task execution. DexMachina~\cite{mandi2025dexmachina}, for example, uses reinforcement learning to track object trajectories from human hand-object demonstrations across different robotic hands. Such methods are powerful, but typically require demonstration data, policy learning, and task-level reward design. DexTwist instead targets online teleoperation of a specific contact-rich primitive: tripod-pinch twisting. Rather than learning a general manipulation policy, it extracts a low-dimensional functional intent signal, namely accumulated twist angle and screw axis direction, from the operator's tripod motion and refines robot joint commands in real time using an explicit geometric objective.

\section{Methodology} \label{sec:methodology}
DexTwist combines an MR-based teleoperation interface with a functional hand retargeting module for contact-rich rotation tasks. The interface provides hand tracking and streamed workspace feedback, while the retargeting module switches between nominal free-space hand motion and functional tripod-pinch twisting. The key component is a screw motion refinement that converts the operator's accumulated twist intent into robot hand commands while stabilizing the robot tripod geometry.

\subsection{Robot Teleoperation System Architecture} \label{sec:teleoperation}

\begin{figure}[t!]
    \centering
    \includegraphics[width=0.99\columnwidth]{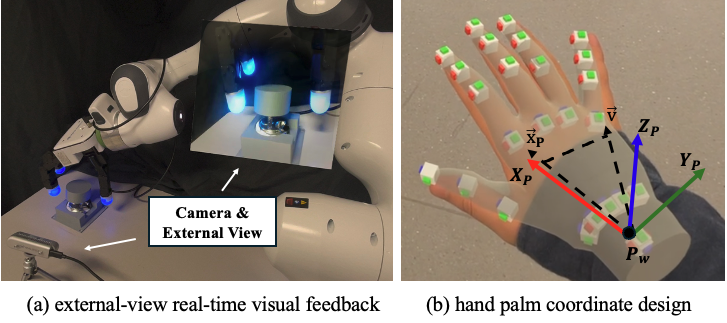}
    \caption{Teleoperation system functionalities demonstration: a) multi-view real-time visual feedback. b) hand palm coordinate construction. }
    \label{fig:2nd figure}
    \vspace{-10pt}
\end{figure}

Following prior work~\cite{cheng2024open,ding2025bunny}, our MR-based teleoperation interface directly leverages the built-in hand pose estimator of an MR headset to obtain the human hand joint configuration $\mathbf{q}_h \in \mathbb{R}^{27}$. Compared to single-view RGB-based approaches~\cite{yan2025learning,qin2023anyteleop}, this design provides improved robustness and consistency, while the internally calibrated sensing pipeline eliminates the need for specialized calibration procedures required by multi-camera or MoCap systems. Real-time visual feedback from external workspace cameras is streamed to the headset for task monitoring. In the current implementation, the headset displays the camera stream as operator feedback rather than using explicit virtual-object overlays. Fig.~\ref{fig:2nd figure}(a) shows a representative workspace view.

Regarding robot motion, the raw hand pose from the MR-headset provides the 3D Cartesian positions of all finger joints and the wrist in the local coordinate system of the headset $\{H\}$. To map these to a robot morphology, we define a local palm frame $\{P\}$ as the primary reference for the control. We designate the wrist keypoint $\mathbf{p}_{w} \in \mathbb{R}^3$ as the origin. To establish a robust orthonormal basis $\mathbf{R}_P = [\mathbf{x}_P, \mathbf{y}_P, \mathbf{z}_P] \in SO(3)$, we utilize the vectors from the wrist to the index knuckle $\mathbf{p}_{k,ind}$ and the pinky knuckle $\mathbf{p}_{k,pnk}$. We define the primary longitudinal axis as:
\begin{equation}
\mathbf{x}_P = \frac{\mathbf{p}_{k,ind} - \mathbf{p}_{w}}{|\mathbf{p}_{k,ind} - \mathbf{p}_{w}|}
\end{equation}

To ensure the $z$-axis represents the palm normal, we compute the normalized cross product with a temporary vector $\mathbf{v} = \mathbf{p}_{k,pnk} - \mathbf{p}_{w}$:
\begin{equation}
\mathbf{z}_P = \frac{\mathbf{x}_P \times \mathbf{v}}{|\mathbf{x}_P \times \mathbf{v}|}
\end{equation}

The basis is completed via $\mathbf{y}_P = \mathbf{z}_P \times \mathbf{x}_P$. The complete pose of the human hand in the headset frame is then represented by the homogeneous transformation matrix $\mathbf{T}_P^H \in SE(3)$:
\begin{equation}
\mathbf{T}_P^H = \begin{bmatrix} \mathbf{x}_P & \mathbf{y}_P & \mathbf{z}_P & \mathbf{p}_w \\ 0 & 0 & 0 & 1 \end{bmatrix}
\end{equation}

To command the robot, we define a static base transformation $\mathbf{T}_H^B$ that aligns the headset's reference frame with the robot's base frame $\{B\}$. The target pose for the robot's end-effector $\mathbf{T}_{cmd}$ is then derived by:
\begin{equation}
\mathbf{T}_{cmd} = \mathbf{T}_H^B \cdot \mathbf{T}_{scale} \cdot \mathbf{T}_P^H
\end{equation}

\noindent where $\mathbf{T}_{scale}$ accounts for the workspace scaling between human and robot reach. This formulation ensures that the incremental changes in $\mathbf{T}_P^H$ are faithfully translated into robot motions, while providing a stable foundation for the task-space twist tracking $\boldsymbol{\mathcal{V}}_{cmd}$ used during functional retargeting. The robot arm adopts a cartesian impedance controller, yielding compliant yet responsive behavior that is well suited for contact-rich manipulation tasks. The dexterous hand is controlled via joint-level commands generated by a hand retargeting module, which maps human hand motion to the corresponding robot hand configuration as stated in next section.

\subsection{Hand Functional Retargeting}\label{sec:functional_retargeting}
Conventional hand retargeting objectives emphasize geometric similarity (e.g., joint angles or fingertip positions). 
For contact-rich twisting, however, small kinematic errors under embodiment mismatch often manifest as tangential slip, axis drift, and loss of tripod stability. DexTwist therefore retargets the twisting function: during a tripod pinch, it estimates the operator’s intended screw axis and accumulated twist in the palm frame, and produces robot hand commands that realize the rotational progress while maintaining a stable tripod geometry.

DexTwist activates twist retargeting when the operator forms a sustained tripod pinch (thumb--index--middle), detected using a hysteresis gate to prevent spurious switching under tracking noise. 
Let $\mathbf{p}_{th},\mathbf{p}_{ind},\mathbf{p}_{mid}\in\mathbb{R}^3$ denote the fingertip positions expressed in the human palm frame $\{P_h\}$. 
% We estimate the instantaneous screw axis from the tripod normal:
We define the screw axis as the normal of the fingertip tripod:
\begin{equation}
\mathbf{a}_h(t) = \mathrm{unit}\!\left((\mathbf{p}_{ind}-\mathbf{p}_{th})\times(\mathbf{p}_{mid}-\mathbf{p}_{th})\right)
\label{eq:human_axis}
\end{equation}
where the axis direction is disambiguated using the palm normal and temporally aligned to avoid sign flips. This definition provides a task frame for estimating the operator's twist intent.
% and disambiguate its sign using the palm normal to avoid axis flips.

Since twisting is frequently executed with ratcheting, DexTwist tracks turning as an accumulated scalar that persists across re-grasps. Specifically, we maintain an accumulated task angle $\theta_{\mathrm{task}}$ that is updated only while tripod contact is active. Using the relative rotation of a tripod-defined tool frame in palm coordinates, the incremental twist about a stabilized axis $\mathbf{a}_{\mathrm{task}}$ is:
\begin{equation}
\label{eq:acc_theta}
\begin{aligned}
    \Delta\theta(t) &= 
\mathbf{a}_{\mathrm{task}}^\top \log\!\Big(\big(\mathbf{R}_{T_h}^{P_h}(t{-}1)\big)^\top \mathbf{R}_{T_h}^{P_h}(t)\Big), \\
&\;\; \theta_{\mathrm{task}} \leftarrow \theta_{\mathrm{task}} + \mathrm{clip}(\Delta\theta(t))
\end{aligned}
\end{equation}

On the robot, we define an analogous tripod tool frame $\{T_r\}$ from the three robot fingertips expressed in the robot palm frame $\{P_r\}$. 
The robot turning progress is computed as the signed rotation of $\{T_r\}$ relative to a reference $\{T_{r,\mathrm{ref}}\}$ latched at tripod activation, projected onto a fixed reference axis $\mathbf{a}_{r,\mathrm{ref}}$:
\begin{equation}
\theta_r(\mathbf{q}) =
\mathbf{a}_{r,\mathrm{ref}}^\top
\log\!\left(\big(\mathbf{R}_{T_{r,\mathrm{ref}}}^{P_r}\big)^\top \mathbf{R}_{T_r}^{P_r}(\mathbf{q})\right)
\label{eq:robot_theta}
\end{equation}

DexTwist then refines the robot hand configuration $\mathbf{q}$ by minimizing a virtual-object objective that tracks turning progress while regularizing tripod stability:
\begin{equation}
\label{eq:func_obj}
\begin{aligned}
\min_{\mathbf{q}}\;\; \mathcal{J}(\mathbf{q}) &=
w_{\mathrm{rot}}\big(\theta_r(\mathbf{q})-\theta_{\mathrm{task}}\big)^2 \\
&\;\; + w_{\mathrm{conn}}\big\lVert \mathbf{e}_r(\mathbf{q})-\mathbf{e}_{r,\mathrm{ref}}\big\rVert^2 \\
&\;\; + w_{\mathrm{axis}}\Big(1-\big(\mathbf{a}_r(\mathbf{q})^\top \mathbf{a}_{r,\mathrm{ref}}\big)^2\Big) \\
&\;\; + w_{\mathrm{pos}}\big\lVert \mathbf{c}_r(\mathbf{q})-\mathbf{c}_{r,\mathrm{ref}}\big\rVert^2 \\
\end{aligned}
\end{equation}
Here $\mathbf{e}_r(\mathbf{q})$ stacks the three pairwise fingertip distances as a closure proxy, $\mathbf{a}_r(\mathbf{q})$ is the instantaneous robot tripod normal, and $\mathbf{c}_r(\mathbf{q})$ is the tripod centroid. The closure, axis, and centroid terms suppress grip opening, axis drift, and translational drift during turning. 
In implementation, Eq.~\eqref{eq:func_obj} is solved at each control cycle as a lightweight residual refinement initialized from the vector-retargeted hand configuration. The optimization variable is the Allegro Hand joint vector $\mathbf{q}$, and the refinement is applied to the tripod-related joints with bounded joint updates. We perform five iterations of finite-difference gradient descent per control cycle, using central differences for gradient estimation and clipped updates for stable real-time execution.

\section{Experiments} \label{sec:experiments}
In this section, we systematically evaluate the effectiveness of the DexTwist framework. We begin by detailing the experimental setup and hardware architecture. Subsequently, we conduct a quantitative comparative analysis against a baseline method, performed in both simulation environments and on physical robot hardware. Finally, we demonstrate the efficacy of the proposed retargeting scheme through qualitative performance in diverse real-world manipulation tasks

\subsection{Experiment Setup} 
\begin{figure}[t!]
    \centering
    \includegraphics[width=0.9\columnwidth]{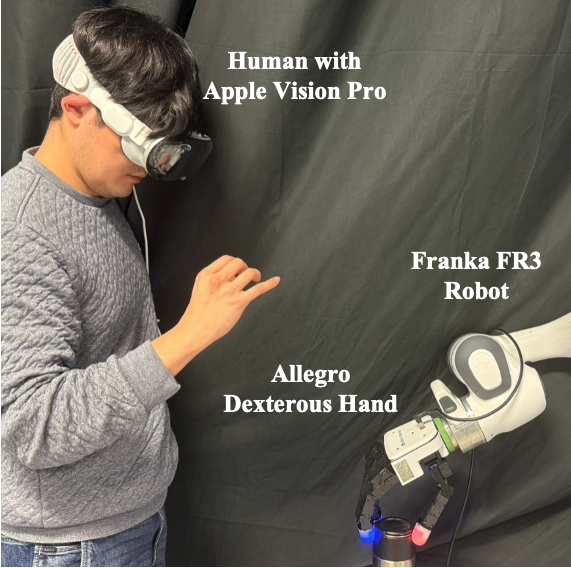}
    \caption{Real-world experiment setup.}
    \label{fig:setup}
    \vspace{-10pt}
\end{figure}

Our experimental setup consists of a human operator wearing an Apple Vision Pro MR headset and a Franka Emika FR3 robotic arm equipped with an Allegro Hand, as shown in Fig.~\ref{fig:setup}. Hand joint configurations and wrist poses are captured using the headset. Based on the estimated wrist pose and hand joint values, relative Cartesian motions guide the robot to an initial configuration for the subsequent twisting task. Real-time camera feedback is streamed to the headset for task monitoring. Before data collection, the operator performed a short familiarization period with the MR interface and the robot response.

\subsection{Baseline Method}
We compare DexTwist against a widely used optimization-based retargeting baseline, \textit{Vector Retargeting}, which retargets human hand motion to the robot by minimizing the discrepancy between task-space vectors, such as palm-to-fingertip and fingertip-to-fingertip vectors~\cite{qin2023anyteleop}. This objective preserves the overall hand shape and relative fingertip layout, and is effective for free-space imitation. However, it does not explicitly encode the functional goal of twist execution (i.e., accumulating rotation about a stable screw axis), and can therefore suffer from axis drift and slip under embodiment mismatch during contact-rich turning.

We use Vector Retargeting as the primary baseline because it directly tests whether replacing kinematic similarity with a twist-specific functional objective improves rotational manipulation. We do not compare against learning-based functional retargeting methods because they require task demonstrations, policy training, and object-centric reward design, whereas DexTwist is designed as an online optimization module for real-time teleoperation.

\subsection{Experimental Results}
\subsubsection{Evaluation Metrics}
We evaluate (i) turning progress tracking and (ii) screw axis stability. Turning progress tracking measures how accurately the robot follows the intended accumulated rotation during twisting. The stability of the tool frame axis measures how well the executed twist remains aligned with a consistent axis, suppressing drift during contact-rich turning.

\smallskip
\noindent\textbf{Turning-angle tracking.}
Let $\theta(t)$ denote the accumulated turning angle of the tripod tool frame expressed in the palm frame (Sec.~\ref{sec:functional_retargeting}). We evaluate tracking performance only when tripod pinch is active, because turning progress is physically meaningful only while the three-finger contact is engaged. Given a ground-truth (GT) turning signal $\theta_{\text{gt}}(t)$ and a method output $\theta_{\text{m}}(t)$, we compute the tracking error:

\begin{equation}
e(t)=\theta_{\text{m}}(t)-\theta_{\text{gt}}(t)
\end{equation}

We additionally report the Root Mean Squared Error (RMSE), Mean Absolute Error (MAE), and Pearson correlation $\rho(\theta_{\text{m}},\theta_{\text{gt}})$:
\begin{equation}
\text{RMSE}=\sqrt{\frac{1}{T}\sum_t e(t)^2}
\end{equation}
\begin{equation}
\text{MAE}=\frac{1}{T}\sum_t |e(t)|
\end{equation}
\begin{equation}
\rho(\theta_{\text{m}},\theta_{\text{gt}})=
\frac{\mathrm{cov}(\theta_{\text{m}},\theta_{\text{gt}})}
{\sigma_{\theta_{\text{m}}}\,\sigma_{\theta_{\text{gt}}}}
\end{equation}

\noindent where $\mathrm{cov}(x,y)$ denotes the covariance and $\sigma$ is the standard deviation.

In screwing, RMSE is sensitive to occasional large errors, such as slippage or re-grasping, whereas MAE summarizes the typical tracking deviation. Pearson correlation captures similarity between the GT turning signal and a method output in the turning progress, increasing during commanded turns and remaining approximately constant during pauses.

\begin{figure}[t!]
    \centering
    \includegraphics[width=0.98\columnwidth]{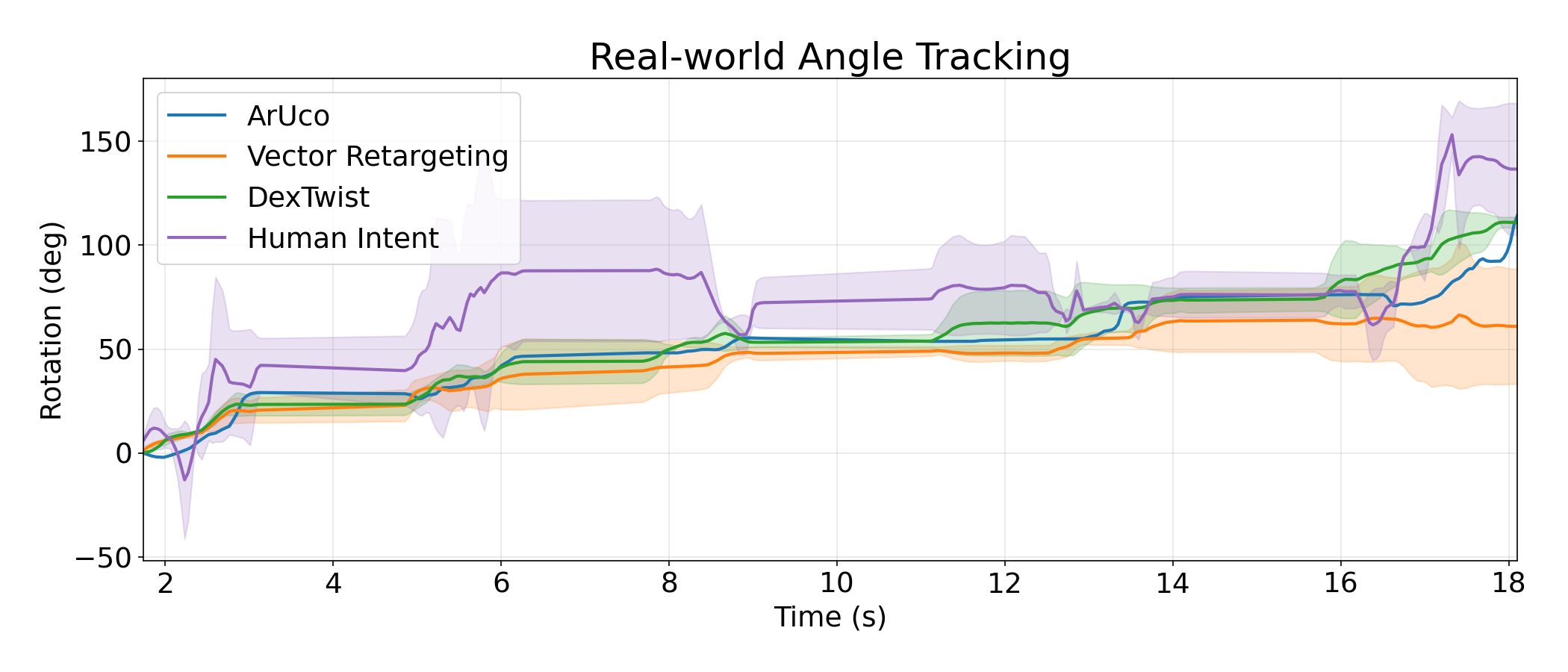}
    \caption{Real-world turning-angle tracking aggregated across trials. \textit{ArUco} denotes object rotation measured from the attached ArUco marker, while \textit{Human Intent} denotes the twist angle estimated from the operator's tracked tripod motion.}
    \label{fig:real_uncertainty}
    \vspace{-10pt}
\end{figure}

\smallskip
\noindent\textbf{Axis deviation.}
To quantify screw axis stability, we compute the instantaneous tripod normal (estimated screw axis) $\mathbf{a}(t)$ from the three robot fingertip positions in the palm frame. Let $\mathbf{a}_{\text{ref}}$ denote the reference axis latched at tripod activation. Axis deviation is defined as:
\begin{equation}
\delta_{\text{axis}}(t) = \arccos\!\left(\left|\mathbf{a}(t)^\top \mathbf{a}_{\text{ref}}\right|\right)
\end{equation}
Lower axis deviation indicates that the estimated robot screw axis remains more consistently aligned with its reference axis during twisting. We report this metric in simulation, where robot fingertip poses are directly available from forward kinematics. In real-world experiments, the ArUco marker provides a direct task-level object rotation measurement. Therefore, real-world evaluation focuses on turning-angle error and temporal correlation.

\begin{table}[b!]
\centering
\caption{Simulation results: tracking error to estimated human intent and screw axis stability.}
\label{tab:sim_metrics}
\setlength{\tabcolsep}{10pt}
\begin{tabular}{lccccc}
\toprule
Method & RMSE($^\circ$) & MAE($^\circ$) &  AxisDev($^\circ$) \\
\midrule
Vector Retargeting & 27.7 & 23.3 & 13.6 \\
DexTwist (Ours)    & \textbf{18.8} & \textbf{15.6} & \textbf{6.1} \\
\bottomrule
\end{tabular}
%\vspace{-8pt}
\end{table}

\subsubsection{Quantitative Result}
We evaluate DexTwist in both simulation and real hardware. Simulation reports tracking to estimated human intent and screw axis consistency from robot fingertip kinematics. In real-world experiments, ArUco-based object rotation is used as the primary task-level reference, while estimated human intent is additionally reported for comparison.

\smallskip
\noindent\textbf{Simulation results.}
In simulation, $\theta_{\text{gt}}(t)$ corresponds to the operator turning intent estimated from the human tripod motion. Table~\ref{tab:sim_metrics} reports angle tracking error to human intent and axis deviation. DexTwist improves turning-angle tracking while reducing axis deviation, confirming that the residual refinement preserves the twist function and stabilizes the tripod geometry. In contrast, Vector Retargeting may satisfy its kinematic objective through tangential fingertip motion under embodiment mismatch rather than rotation about a consistent screw axis, leading to degraded accumulated turning progress.

\begin{table}[b!]
\centering
\caption{Real-world angle tracking results.}
\label{tab:real_metrics}
\setlength{\tabcolsep}{12pt}
\begin{tabular}{lccc}
\toprule
\multicolumn{4}{c}{\textbf{Reference = ArUco}} \\
\midrule
Method & RMSE ($^\circ$) & MAE ($^\circ$) & Corr \\
\midrule
Vector Retargeting & 23.2 & 15.6 & 0.83 \\
DexTwist (Ours)    & \textbf{14.3} & \textbf{11.9} & \textbf{0.96} \\
\midrule
\multicolumn{4}{c}{\textbf{Reference = Human Intent}} \\
\midrule
Method & RMSE ($^\circ$) & MAE ($^\circ$) & Corr \\
\midrule
Vector Retargeting & 40.3 & 32.8 & 0.60 \\
DexTwist (Ours)    & \textbf{33.7} & \textbf{25.9} & \textbf{0.67} \\
\bottomrule
\end{tabular}
%\vspace{-8pt}
\end{table}

\begin{figure}[t!]
   \centering
  \includegraphics[width=0.98\columnwidth]{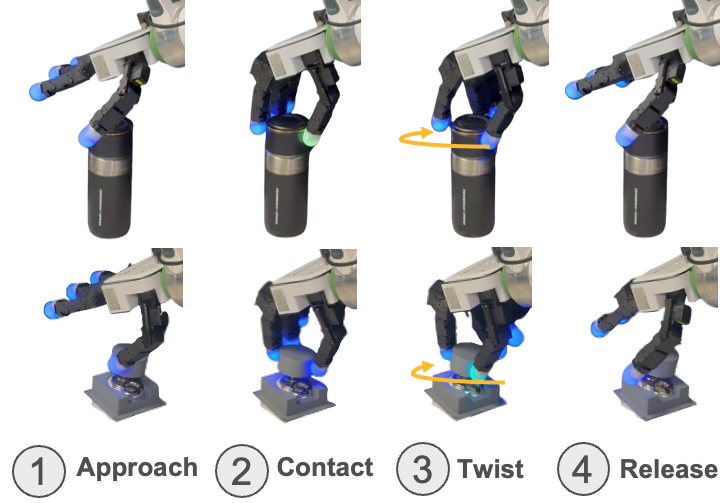}
  \caption{Real-world demonstrations for the cap screwing and key unlocking with the proposed DexTwist framework.}
  \label{fig:real_demo}
  \vspace{-10pt}
\end{figure}

\smallskip
\noindent\textbf{Real-world results.}
In real-world experiments, all trials are conducted by one operator after familiarization within 10 hours usage. Each method was tested over five trials using the same object setup and comparable initial configurations. An ArUco marker was rigidly attached to the manipulated object to estimate object rotation, yielding $\theta_{\text{aruco}}(t)$.

Fig.~\ref{fig:real_uncertainty} shows the turning-angle trajectories aggregated across five trials, and Table~\ref{tab:real_metrics} reports the corresponding metrics. DexTwist achieves lower tracking error and higher temporal agreement than Vector Retargeting, indicating more reliable accumulation of turning progress. Late-phase errors can increase due to partial fingertip occlusion, re-grasping, and accumulated orientation mismatch, which bias the visual estimate of human intent. Although DexTwist tracks Human Intent as its reference, the stability regularizers can yield closer alignment with measured object rotation than with the noisy intent estimate.

\subsubsection{Real-world Demonstration}
Fig.~\ref{fig:real_demo} illustrates representative real-world tasks enabled by DexTwist. The system supports contact-rich rotational manipulation under tripod pinch, including continuous turning and regrasping while maintaining stable fingertip contacts. We demonstrate two twist interactions: cap screwing and key unlocking. Across all tasks, DexTwist preserves turning progress through re-grasp transitions and executes rotation about a consistent axis, enabling practical twist motions without task-specific object models.

\section{Discussion \& Conclusion} \label{sec:conclusion}

In this work, we presented DexTwist, an MR-based dexterous teleoperation framework for contact-rich rotational tasks such as cap opening, key turning, and bolt screwing. Under human-robot embodiment mismatch, conventional kinematic retargeting can preserve fingertip pose similarity while failing to preserve the intended manipulation function. DexTwist addresses this by estimating the operator's tripod-defined screw axis and accumulated turning progress, then refining robot hand commands with a virtual-object objective defined on the robot tripod geometry. Simulation and real-world experiments show improved turning-angle tracking over Vector Retargeting, together with reduced screw axis deviation in simulation.

However, there are several limitations. The current formulation focuses on tripod-pinch twisting, where three non-collinear fingertip contacts define the screw axis and accumulated twist. This provides a stable representation for residual refinement, but does not cover all rotational manipulation strategies. Depending on object size, required torque, and task context, two/four-finger pinching, cylindrical grasps, power grasps, or tool-specific grasps may be more appropriate. Extending DexTwist to these cases would require grasp-dependent functional representations estimated from contact geometry, object pose, or tactile feedback. The system also relies on visual hand tracking, so severe self-occlusion can bias the inferred twist intent. Incorporating tactile or contact-state sensing could improve robustness to slip and unreliable visual updates.

From a societal and economic perspective, reliable dexterous teleoperation can support remote operation in hazardous, costly, or ergonomically demanding environments, including maintenance, inspection, disaster response, and industrial servicing. By improving contact-rich teleoperation reliability, DexTwist may reduce the need for direct human presence in unsafe environments and support scalable collection of high-quality manipulation demonstrations for robot learning.

\addtolength{\textheight}{-2cm}   % This command serves to balance the column lengths

%%%%%%%%%%%%%%%%%%%%%%%%%%%%%%%%%%%%%%%%%%%%%%%%%%%%%%%%%%%%%%%%%%%%%%%%%%%%%%%%

\bibliographystyle{IEEEtran}
\bibliography{references}

\end{document}